\documentclass[letterpaper, 10 pt, conference]{ieeeconf}  % Comment this line out if you need a4paper

\IEEEoverridecommandlockouts                              % This command is only needed if
\makeatletter
\newcommand*{\rom}[1]{\expandafter\@slowromancap\romannumeral #1@}
\makeatother

\usepackage[bindingoffset=0.2in,%
            left=0.75in,right=0.75in,top=0.75in,bottom=0.75in,%
            footskip=.25in]{geometry}
\usepackage{graphics} % for pdf, bitmapped graphics files
\usepackage{epsfig} % for postscript graphics files
\usepackage{epstopdf}
\graphicspath{ {figs} }
\usepackage{graphicx}
\usepackage{mwe}
\usepackage{caption}
\usepackage{xcolor}
\usepackage{subcaption}
\usepackage{lipsum}
\usepackage{float}
\usepackage{todonotes}
\usepackage{color,soul}
\usepackage{hyperref}
\usepackage{multirow}
\usepackage{algorithm}
\usepackage{algorithmic}
\usepackage{cuted}

\usepackage{amsmath} % assumes amsmath package installed

\newcommand{\argminE}{\mathop{\mathrm{argmin}}}
\usepackage[scientific-notation=true]{siunitx}
\usepackage{amssymb}  % assumes amsmath package installed
\usepackage{array}
\newcolumntype{x}[1]{>{\centering\arraybackslash\hspace{0pt}}p{#1}}

\newcommand{\bx}{\mathbf{x}}

\newcommand{\by}{\mathbf{y}}

\newcommand{\bD}{\mathbf{D}}
\newcommand{\bE}{\mathbf{E}}
\newcommand{\bsigma}{\mathbf{\sigma}}

\newcommand{\cX}{\mathcal{X}}

\definecolor{dblue}{rgb}{0,0,0.7}

\title{\LARGE \bf
		Visual-based Autonomous Driving Deployment from a Stochastic and Uncertainty-aware Perspective
}

\author{
Lei Tai \ \ \
Peng Yun \ \ \
Yuying Chen \ \ \
Congcong Liu \ \ \
Haoyang Ye \ \ \
Ming Liu \ \ \
\thanks{This work was supported by the National Natural Science Foundation of China (Grant No. U1713211), the Research Grant Council of Hong Kong SAR Government, China, under Project No. 11210017, and No. 21202816, and Shenzhen Science, Technology and Innovation Comission (SZSTI) JCYJ20160428154842603, awarded to Prof. Ming Liu.}
\thanks{All authors are with The Hong Kong University of Science and Technology (email: \{ltai, pyun, ychenco, cliubh, hyeab, eelium\}@ust.hk).}
}

\begin{document}
\maketitle

%%%%%%%%%%%%%%%%%%%%%%%%%%%%%%%%%%%%%
\begin{strip}\centering
  \vspace{-2.2cm}
  \includegraphics[width=\textwidth]{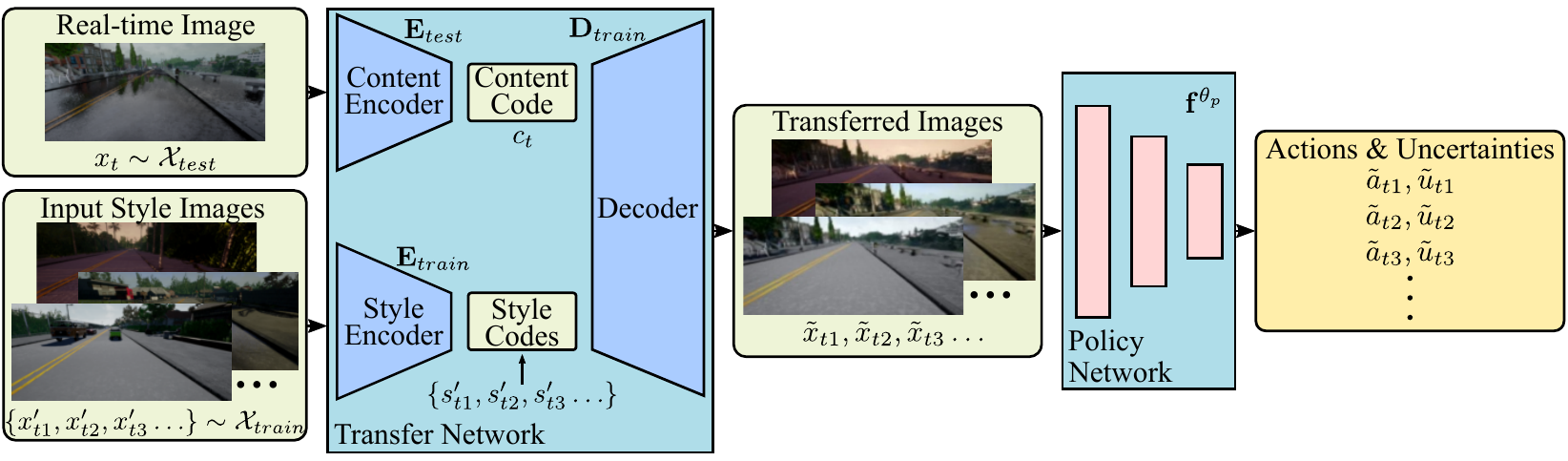}
  \captionof{figure}{
    The proposed end-to-end visual navigation deployment pipeline. When applying the end-to-end visual-based driving policy in real-time, an image from the mounted sensor is first translated to the training domain under various conditions through a stochastic generator.
    An uncertainty-aware imitation learning policy forward processes all the transferred images to output actions with uncertainties.
    The uncertainty can be used to refer to the most certain action.
		%for deploy.
  % \hl{skip the structure of the pollicy network, velocity and else}
  \label{fig:real_time_pipeline}}
  % \vskip -0.4cm
\end{strip}
%%%%%%%%%%%%%%%%%%%%%%%%%%%%%%%%%%%%%

\thispagestyle{empty}
\pagestyle{empty}

%%%%%%%%%%%%%%%%%%%%%%%%%%%%%%%%%%%%%%%%%%%%%%%%%%%%%%%%%%%%%%%%%%%%%%%%%%%%%%%%
\begin{abstract}
  End-to-end visual-based imitation learning has been widely applied in autonomous driving. When deploying the trained visual-based driving policy, a deterministic command is usually directly applied without considering the uncertainty of the input data. Such kind of policies may bring dramatical damage when applied in the real world.
  In this paper, we follow the recent \textit{real-to-sim} pipeline by translating the testing world image back to the training domain when using the trained policy. In the translating process, a stochastic generator is used to generate various images stylized under the training domain randomly or directionally. Based on those translated images, the trained uncertainty-aware imitation learning policy would output both the predicted action and the data uncertainty motivated by the \textit{aleatoric} loss function. Through the uncertainty-aware imitation learning policy, we can easily choose the safest one with the lowest uncertainty among the generated images. Experiments in the \textit{Carla} navigation benchmark show that our strategy outperforms previous methods, especially in dynamic environments.

  % \hl{provide explanation}

% % Problem is still challenging
% Sucessfully deploy the trained driving poliy from simulation to the real world is challenging.
% % Real to virtual make sense
%
% % In this pipeline we
% Uncertainty Aware
% stocastic style transfer and uncertainty aware policy network training
% naturally can be solved through this pipeline
%
% % experiment result
% we achieve the benchmark higher than before.
% (do we need to do real world experiments?)

\end{abstract}

%%%%%%%%%%%%%%%%%%%%%%%%%%%%%%%%%%%%%%%%%%%%%%%%%%%%%%%%%%%%%%%%%%%%%%%%%%%%%%%%
\section{Introduction}
\label{sec:introduction}

End-to-end visual-based driving has received various interest from both the deep reinforcement learning \cite{dosovitskiy2017carla, zhang2019vrgoggles} and imitation learning \cite{Codevilla2018} perspectives. In this paper, we mainly consider visual-based imitation learning, where a model is trained to guide a vehicle behaving similarly to a human demonstrator based on visual information. As a model-free method, raw visual information and other related measurements are taken as the input of a deep model, which is commonly a deep convolutional neural network (CNN) model. The deep model then outputs control commands directly, like steering and acceleration. It has been successfully applied in both indoor navigation \cite{tai2016deep} and outdoor autonomous driving \cite{Codevilla2018}.

Though learning-based methods have achieved many breakthroughs for autonomous driving and mobile robot navigation, the uncertainty is rarely considered when deploying the trained policy. However, uncertainty is critical for robotics decision making. Unlike other pure perception scenarios, where higher uncertainty of the prediction may influence the accuracy of a segmentation mask or output an incorrect classification result, the uncertain decision in autonomous driving endangers the safety of vehicles or even human lives. Thus, we should not always assume that the output of the deep model is accurate. Knowing what a model does not understand is an essential part, especially for autonomous driving under dynamic environments and interacting with pedestrians and vehicles.

When arranging the policy in the testing world, like the real world, a common pipeline is translating the visual input from the real world back to the training simulation environment \cite{zhang2019vrgoggles, yang2018eccv} through a generative adversarial network (GAN). Most of the previous works have focused on image-to-image transfer through a deterministic generator.
However, the imitation learning policy is usually trained in a multi-domain environment with various conditions for better generalization. Thus, for a deterministic translation, the problem is which training scenario should we transfer the real-world image to. In this paper, we extend this pipeline to generate various translated images with training data styles through multimodal cross-domain mapping. To generate the transferred images, we can randomly sample style codes from a normal distribution or directionally encode the provided style images from the training domain. The content code is extracted from the real-world image collected from the mounted sensor in real-time. A decoder would take the content code and style codes as input to generate various stylized images.

Naturally, we can predict the actions and uncertainties of all the translated images through the proposed uncertainty-aware imitation learning network. Among the generated images, the most certain one will be considered to deploy to the agent.

We list the main contributions of our work as follows:
\begin{itemize}
  \item We transfer the real driving image back to diverse images stylized under the familiar training environment through a stochastic generator so that the decision is made through multiple alternate options.
  \item The uncertainty-aware imitation learning network provides a considerable way to make driving decisions, which improves the safety of autonomous driving, especially in dynamic environments.
  \item We explain the aleatoric uncertainty from the view of the noisily labelled data samples.
\end{itemize}

\section{Related Works}
\label{sec:relatedWorks}
In this section, we review related works in end-to-end driving, uncertainty-aware decision making and visual domain adaptation.

\subsection{End-to-end Driving}
\label{sec:end2end-driving}

Traditional visual-based strategies in autonomous driving and robot navigation, traditional methods firstly recognise relevant objects from visual inputs, including pedestrians, traffic lights, lanes, and cars.
That information is considered to make the final driving decisions based on manually designed rules \cite{Chen_2015_ICCV}.
Recently, benefits from the excellent approximation ability of deep neural networks, end-to-end methods have become more and more popular in vision-based navigation.

Tai \textit{et al.} \cite{tai2016deep} used deep convolutional neural networks to map depth images to steering commands so that the agent can make meaningful decisions like a human demonstrator in an indoor corridor environment. A similar framework was also successfully applied in a forest trail scenario to navigate a flying platform for obstacle avoidance \cite{giusti2016machine}.
They also considered softly combining all the discrete commands based on the weighted outputs of the softmax structure. Codevilla \textit{et al.} \cite{Codevilla2018} designed a deep structure with multiple branches for end-to-end driving through imitation learning. Based on the high-level commands from the global path planner, outputs from the specific branch are applied to the mobile agent.

Reinforcement learning (RL) algorithms also show surprising effects in end-to-end navigation. Zhang \textit{et al.} \cite{zhang2017irosdeep} explored the target-arriving ability of a mobile robot through RL based on a single depth image. Their policy can also quickly adapt to new situations through successor features. For autonomous driving, RL algorithms are also considered to train an intelligent agent through interaction with simulated environments like \textit{Carla} \cite{dosovitskiy2017carla}. Liang \textit{et al.} \cite{Liang_2018_ECCV} used the model weight trained through imitation learning as the initialization of their reinforcement learning policy, while Tai \textit{et al.} \cite{tai2018social} proposed to solve the socially compliant navigation problem through inverse RL. However, all of the methods above directly deploy the learned policy on related platforms. None of them considers the uncertainty of the decision.

\subsection{Uncertainty in learning-based decision making}
\label{sec:uncertainty-deep-learning}

The uncertainty in deep learning is derived from the \textit{Bayesian deep learning} \cite{gal2016uncertainty} approaches, where \textit{aleatoric} uncertainty and \textit{epistemic} uncertainty are extracted through specific learning structures \cite{kendall2017uncertainties}. Recently, computer vision researchers have started to leverage those uncertainties on related applications, like balancing the weight of different loss items for multi-task visual perception \cite{kendall2018multi}. The uncertainty estimation helps the deep \textit{Bayesian} models to achieve state-of-the-art results on various computer vision benchmarks, including semantic segmentation and depth regression.

In terms of decision making in robotics, Kahn \textit{et al.} \cite{kahn2017uncertainty} proposed an uncertainty-aware model-based reinforcement learning method to update the confidence for a specific obstacle iteratively. During the training phase, the agent behaves more carefully in unfamiliar scenarios at the beginning. Based on this work, Lutjens \textit{et al.} \cite{lutjens2018safe} explored more complex pedestrian-rich environments. The uncertainty was further considered for the exploitation and exploration balance in their implementation \cite{lutjens2018safe}.
Henaff \textit{et al.} \cite{henaff2019model} focused on the out-of-distribution data where an uncertainty cost was used to represent the divergence of the test sample from the training states. However, all of the methods above follow a pipeline using multiple stochastic forward passes through \textit{Dropout} to estimate the \textit{epistemic} uncertainty \cite{kendall2017uncertainties}. The time-consuming computation potentially limits these methods in scenarios which ask for real-time deployment ability.

A highly related work is the work of Choi \textit{et al.} \cite{choi2018uncertain}. They proposed a novel uncertainty estimation method where a single feedforward is enough for uncertainty acquisition. However, they only tested their method in the state space. In this paper, we aim to tackle a much more difficult visual-based navigation problem.

\subsection{Visual domain adaptation}
\label{sec:visual-domain}

For a policy trained in simulated environments or based on datasets collected from simulated environments, the gap with the testing world (e.g. the real world) is always an essential problem. In the following, we mainly review the policy transferring methods through image translation.

One probable solution is the so-called \textit{sim-to-real}, where synthetic images are translated to the realistic domain \cite{pan2017virtual}. With an additional adaptation step for each training iteration, the whole training-deployment procedure is inevitably slowed down.

Another direction is \textit{real-to-sim}, where real-world images are translated back to simulated environments.
Zhang \textit{et al.} \cite{zhang2019vrgoggles} extended the \textit{CycleGAN} \cite{zhu2017unpaired} framework with a \textit{shift loss}, which improves the consistency of the generated image streams. They achieved great improvements on the \textit{Carla} \cite{dosovitskiy2017carla} navigation benchmark.
Muller \textit{et al.} \cite{mueller18corl} firstly perceived a real-world RGB image as a segmentation mask which is used to generate path points through a learned policy network. %\hl{however, it is not end-to-end}.

For the purely unsupervised image-to-image translation problem, unlike the previous deterministic translation model \cite{zhu2017unpaired}, multimodel mapping has received lots of attention from computer vision researchers \cite{Huang_2018_ECCV, lee2018diverse, almahairi2018augmented}.
Their goal is translating an image from the source domain to a conditional distribution of the related image in the target domain. This is a naturally applicable method for a robotic task because the training domain of the policy networks always contains data collected from various conditions (e.g. different weathers \cite{Codevilla2018, zhang2019vrgoggles}) for better generalization ability.

\section{An explanation of \textit{Aleatoric} uncertainty}
\label{sec:uncertainty}

As mentioned before, there are two types of uncertainty in deep learning as introduced in \cite{gal2016uncertainty} and \cite{kendall2017uncertainties}, the \textit{aleatoric} uncertainty and the \textit{epistemic} uncertainty. The \textit{epistemic} uncertainty is the model uncertainty, which can be reduced by adding enough data.
However, it is commonly realized through stochastic \textit{Dropout} forward passes, which cost too much time to be applied in real time. In this paper, we mainly consider the \textit{aleatoric} uncertainty, the data uncertainty.

Following the heteroscedastic \textit{aleatoric} uncertainty setup in \cite{kendall2017uncertainties}, a regression task can be represented as
\begin{align}
[\tilde{\by}, \tilde{\sigma}] &= \mathbf{f}^{\theta}(\bx), \\
\mathcal{L}(\theta) &= \frac{1}{2} \frac{\| \by - \tilde\by \|^2}{\tilde\sigma^2} + \frac{1}{2} \log \tilde\sigma^{2}.
\end{align}
Here, $\bx$ is the input data, $\by$ and $\tilde\by$ are the groundtruth regression target and the predicted result. $\tilde\sigma$ is another output of the model and can represent the standard variance of the data $\bx$, and $\theta$ is the model weight of the regression model.

We provide an explanation for $\tilde\sigma$ to show why it can represent the standard variance or the uncertainty of $\bx$. Suppose that there is a subset $\Psi$ of the training dataset, $\{{\bx_i, \by_i}\}_{i=1}^{N_\psi}$ with size $N_\psi$.
For the prediction and the uncertainty,
$\{\tilde\by_i, \tilde\bsigma_i: [\tilde\by_i, \tilde\bsigma_i] = \mathbf{f}^{\theta_\psi}(\bx_i)\}_{i=1}^{N_\psi}$. The optimization target of this subset is
\begin{align}
\min \mathcal{L}(\theta_\psi) = \min \sum_{i}^{N_\psi} \frac{1}{2} \frac{\| \by_i - \tilde\by_i \|^2}{\tilde\sigma_i^2} + \frac{1}{2} \log \tilde\bsigma_i^{2}.
\end{align}
$\theta_\psi$ is the model weight to optimize for this subset $\Psi$. Assume that all the $\bx_i$ in this subset are exactly the same, as $\bx_\psi$. Because of the limitations of human labelling, they may be labelled with conflicting ground truths (like the noise labels around object boundaries in \cite{kendall2017uncertainties}).
Then, the model will output the same prediction $\tilde\by_\psi$ and uncertatinty $\tilde\sigma_\psi$ for all of $\{\bx_i\}_{i=1}^{N_\psi}$ as $[\tilde\by_\psi, \tilde\bsigma_\psi] = \mathbf{f}^{\theta_\psi}(\bx_\psi)$.
The minimization target turns to
\begin{align}
\min \mathcal{L}(\theta_\psi) = \min \sum_{i}^{N_\psi} \frac{1}{2} \frac{\| \by_i - \tilde\by_\psi \|^2}{\tilde\sigma_\psi^2} + \frac{1}{2} \log \tilde\bsigma_\psi^{2}.
\end{align}
Considering that $\tilde\by_\psi$ and $\tilde\sigma_\psi$ are conditionally independent on $\bx_\psi$, $\tilde{\sigma}_\psi$ can be derived through the first-order derivative as
\begin{align}
  \frac{\partial \mathcal{L}(\theta_\psi)}{\partial \tilde\sigma_\psi}
  = & \sum_{i}^{N_{\psi}} -  \frac{\| \by_i - \tilde\by_\psi \|^2}{\tilde\sigma_\psi^3}
  +  \frac{1}{\tilde\bsigma_\psi} \nonumber \\
  = & \sum_{i}^{N_{\psi}} -  \frac{\| \by_i - \tilde\by_\psi \|^2}{\tilde\sigma_\psi^3}
  +  \frac{1}{\tilde\bsigma_\psi} =0 \nonumber\\
  \Rightarrow & \sum_{i}^{N_{\psi}} \frac{\| \by_i - \tilde\by_\psi \|^2}{\tilde\sigma_\psi^3}
  = \frac{N_{\psi}}{\tilde\bsigma_\psi} \nonumber\\
  & \frac{\sum_{i}^{N_{\psi}} \| \by_i - \tilde\by_\psi \|^2}{N_{\psi}}
  = \tilde\bsigma_\psi^2.
\end{align}
For the model $\mathbf{f}^{\theta_\psi}$, it makes sense to output $\by_\psi$ as the mean of $\{\by_i\}_{i=1}^{N_\psi}$. And that is why $\tilde\sigma_\psi^2$, as the prediction variance of the $\{\by_i\}_{i=1}^{N_\psi}$, can be regarded as the uncertainty of $\{{\bx_i, \by_i}\}_{i=1}^{N_\psi}$.
For a decision-making task, even though at some point, the model cannot predict a good enough command, it should know this prediction is uncertain but not directly deploy it.

\section{Implementations}
\label{sec:methods}

\subsection{Carla navigation dataset and benchmark}
\label{sec:carla_benchmark}

As mentioned in \cite{zhang2019vrgoggles}, it is difficult to evaluate the autonomous driving policy under a common benchmark in the real world. Thus,
we use the \textit{Carla} driving dataset\footnote{https://github.com/carla-simulator/imitation-learning} to train the visual-based navigation policy. Then for the evaluation, we can naturally deploy it through the \textit{Carla} navigation benchmark \cite{dosovitskiy2017carla, Codevilla2018} under an unseen extreme weather condition.
The distribution of the \textit{Carla} dataset \cite{dosovitskiy2017carla} and the benchmark details are available in \cite{zhang2018vrsup}.

\begin{figure}[t]
    \centering
    \begin{subfigure}{0.49\columnwidth}
      \includegraphics[width=\columnwidth]{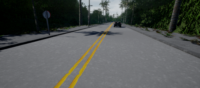}
      \caption{\textit{daytime}}
      \label{fig:carla-a}
    \end{subfigure}
    \begin{subfigure}{0.49\columnwidth}
      \includegraphics[width=\columnwidth]{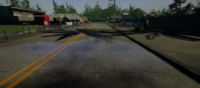}
      \caption{\textit{daytime after rain}}
      \label{fig:carla-b}
    \end{subfigure}
    \begin{subfigure}{0.49\columnwidth}
      \includegraphics[width=\columnwidth]{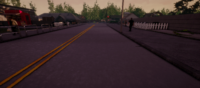}
      \caption{\textit{clear sunset}}
      \label{fig:carla-c}
    \end{subfigure}
    \begin{subfigure}{0.49\columnwidth}
      \includegraphics[width=\columnwidth]{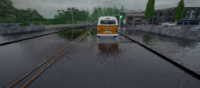}
      \caption{\textit{daytime hard rain}}
      \label{fig:carla-d}
    \end{subfigure}
    \caption{\textit{Carla} weather conditions considered in this paper: training conditions including (a) \textit{daytime}, (b) \textit{daytime after rain} and (c) \textit{clear sunset} and the testing condition (d) \textit{daytime hard rain}.}
    \label{fig:carla-weathers}
\end{figure}

The collected expert dataset of \textit{Carla} includes four different weather conditions (\textit{daytime, daytime after rain, clear sunset and daytime hard rain}). The original experiments in \cite{dosovitskiy2017carla} tested its policy under \textit{cloudy daytime} and \textit{soft rain at sunset}. However, considering these two weathers are not available in the provided dataset for domain adaptation, we resplit the \textit{Carla} driving dataset into a training domain (\textit{daytime, daytime after rain, clear sunset}) and testing domain (\textit{daytime hard rain}), as shown in Fig. \ref{fig:carla-weathers}, following the setup in \cite{zhang2019vrgoggles}.
The vehicle speed, ground truth actions and related measurements are also provided by the dataset \cite{dosovitskiy2017carla} and considered by our policy model.

% \hl{before that all real to sim}
The final testing environment under \textit{daytime hard rain} is very challenging. We believe that the difficulty in deploying the policy through visual domain transformation from the testing domain to the training domain (\textit{test-to-train}) in this paper can be regarded as comparable to the previous \textit{real-to-sim} experiments \cite{zhang2019vrgoggles, yang2018eccv}.

% \hl{regard hard rain as the "real" environment, it is the extreme weather wich is comparable as to real world}
% \hl{25 under dynamic}
% \hl{Notice that we use real domain and sim domain}

\subsection{Uncertainty-aware Imitation Learning}
\label{sec:uail}

\begin{figure}[t!]
    \centering
    \includegraphics[width=\columnwidth]{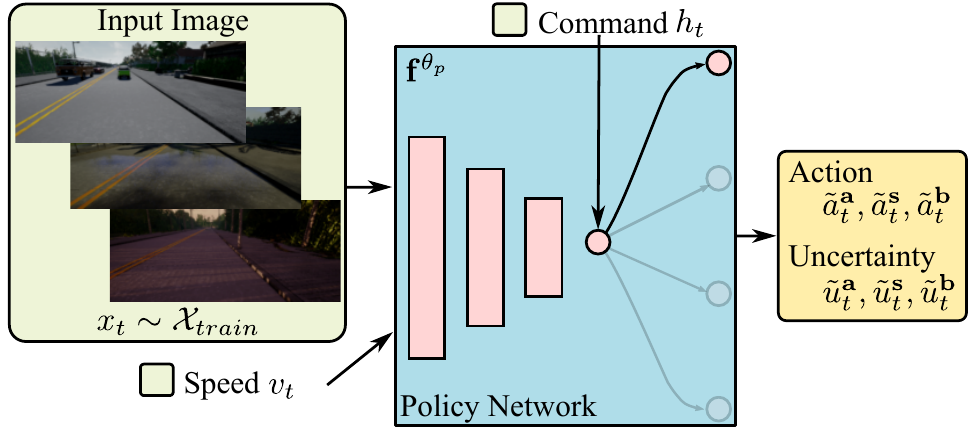}
    \caption{The uncertainty-aware imitation learning pipeline of this paper. Based on the branched structure of the conditional imitation learning \cite{Codevilla2018}, a first-person-view image and the related velocity are taken as the input of the network. The final output is from the branch decided by the corresponding high-level command. The network also further generates uncertainties matching each output. The loss function is described in Section \ref{sec:uail}.}
    \label{fig:policy_networks}
\end{figure}

We first introduce the framework of the policy network, which is the proposed uncertainty-aware imitation learning network, as shown in Fig. \ref{fig:policy_networks}. The backbone of our policy network is based on the conditional imitation learning network \cite{Codevilla2018} for visual-based navigation.

For this framework, the training dataset includes all three kinds of weather in the training domain mentioned in Section \ref{sec:carla_benchmark}.
In each forward step, an RGB Image $x_t$ from the training dataset and the related vehicle's speed $v_t$ are taken as the input to the network. The extracted features are passed to four different branches with the same structures. A high-level command $h_t$ (\textit{straight, left, right, follow line})
from the global path planner decides that output from which the branch will be chosen as the final prediction. The output consists of the predicted action $\tilde{a}_t$ and its estimated uncertainty $\tilde\sigma_t$.
In practice, as in \cite{kendall2017uncertainties}, we let the netowrk predict the log variance $\tilde{u}_t:=\log \tilde\sigma_t^2$.
The action $\tilde{a}_t$ is actually a vector including
accelaration $\tilde{a}_t^\mathbf{a}$,
steering $\tilde{a}_t^\mathbf{s}$, and
braking $\tilde{a}_t^\mathbf{b}$, with their related uncertainties $\tilde{u}_t^\mathbf{a}$, $\tilde{u}_t^\mathbf{s}$,
and $\tilde{u}_t^\mathbf{b}$ repectively.

We use ${\theta}_p$ to represent the weight of the policy network. $a_t^\mathbf{a}, a_t^\mathbf{s}$ and $a_t^{\mathbf{b}}$ represent the groundtruth actions from the collected dataset.
The policy prediction process $\mathbf{f}^{\theta_p}$ and our uncertainty-aware loss function $\mathcal{L}_p$ are as follows:
\begin{align}
[\tilde{a}_t^\mathbf{a}, \tilde{a}_t^\mathbf{s}, \tilde{a}_t^\mathbf{b},
 \tilde{u}_t^\mathbf{a}, \tilde{u}_t^\mathbf{s}, \tilde{u}_t^\mathbf{b}]
                        &= \mathbf{f}^{\theta_p}(x_t, v_t, h_t)
\end{align}
\begin{align}
\mathcal{L}_p(\theta_p) &= \frac{1}{2} \exp(- \tilde{u}_t^\mathbf{a}) \| a_t^\mathbf{a} - \tilde{a}_t^\mathbf{a} \|^2 + \frac{1}{2} \tilde{u}_t^\mathbf{a}  \nonumber \\
                        &+ \frac{1}{2} \exp(- \tilde{u}_t^\mathbf{s}) \| a_t^\mathbf{s} - \tilde{a}_t^\mathbf{s} \|^2 + \frac{1}{2} \tilde{u}_t^\mathbf{s}  \nonumber \\
                        &+ \frac{1}{2} \exp(- \tilde{u}_t^\mathbf{b}) \| a_t^\mathbf{b} - \tilde{a}_t^\mathbf{b} \|^2 + \frac{1}{2} \tilde{u}_t^\mathbf{b}\text{.}
\end{align}

\subsection{Stochastic real(test)-to-sim(train) Transformation}
\label{sec:stochastic_transfer}

\begin{figure}[t!]
    \centering
    % \begin{subfigure}{0.49\columnwidth}
      \includegraphics[width=0.6\columnwidth]{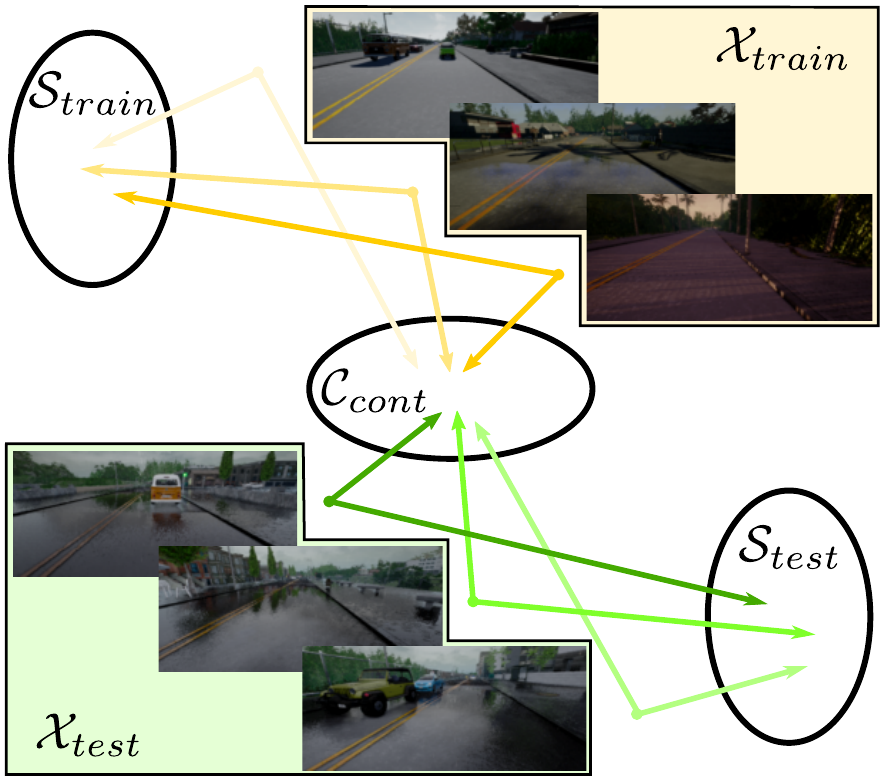}
    \caption{
    The stochastic visual domain transformation structure following \cite{Huang_2018_ECCV}.
    The training domain $\mathcal{X}_{train}$ consists of three weather conditions and the testing domain $\mathcal{X}_{test}$ is under a specific weather condition. These two domains share the same content space and maintain their own style space. The setting pipeline is explained in Section \ref{sec:stochastic_transfer}.
    }
    \label{fig:stochastic_gan}
\end{figure}

For the unsupervised \textit{real-to-sim} pipeline, previous works \cite{yang2018eccv, zhang2019vrgoggles} are based on a deterministic structure like \textit{CycleGAN} \cite{zhu2017unpaired}.
In this paper, we mainly consider stochastic multimodel translation \cite{Huang_2018_ECCV} through GAN.
The training domain, with three different kinds of weather, is represented as $\mathcal{X}_{train}$, and the testing domain, with a single weather condition, is represented as $\mathcal{X}_{test}$, as shown in Fig. \ref{fig:stochastic_gan}. These two domains are supposed to maintain their distinguishable style space ($\mathcal{S}_{test}$ and $\mathcal{S}_{train}$) but share a common content space $\mathcal{C}_{cont}$.
The stochastic model contains an encoder ($\bE_{test}, \bE_{train}$) and a decoder ($\bD_{test}, \bD_{train}$) for each of the domains.
The training procedure follows the setup in \cite{Huang_2018_ECCV}. For example, an image $x_1 \sim \mathcal{X}_{train}$ sampled from the training domain can be encoded to its style code $s_1$ and content code $c_1$ by the training domain encoder $\bE_{train}$. The training domain decoder can also combine these two codes to generate $\tilde{x}_1$ as the reconstruction of $x_1$ as follows:
\begin{align}
      [c_1, s_1] & = \bE_{train}(x_1) \\
      \tilde{x}_1 & = \bD_{train}(c_1, s_1)\text{.}
\end{align}
For the cross-domain translation, the testing domain decoder $\bD_{test}$ combines a random style code $\hat{s}_2$ from the testing domain and the content code $c_1$ to generate the translated image $\tilde{x}_{1 \rightarrow 2}$. The testing domain encoder $\bE_{test}$ takes this translated image as input and generates $\tilde{c}_1$ and $\tilde{s}_2$ as the reconstruction of $c_1$ and $\hat{s}_2$ as follows:
\begin{align}
      \tilde{x}_{1 \rightarrow 2} & = \bD_{test}(c_1, \hat{s}_2) \\
      [\tilde{c}_1, \tilde{s}_2] & = \bE_{test}(\tilde{x}_{1 \rightarrow 2}) \text{.}
\end{align}
($\hat{s}_2, \tilde{s}_2$), ($c_1, \tilde{c}_1$), and ($x_1$, $\tilde{x}_1$) are constrained by L1 loss.
 A discriminator of the GAN structure is used to distinguish $\tilde{x}_{1 \rightarrow 2}$ from the original testing domain images.
 We skip the reconstruction of the testing domain image and the translation procedure from the testing domain to the training domain.
The content code is a 2D matrix with a size corresponding to the input image. The style code is a vector with eight individually sampled numbers from the normal distribution. Note that the whole pipeline is unsupervised. Images from the two domains do not need to be paired for the training.

\subsection{Deloyment phase}
\label{sec:forward_pipeline}

In the deployment phase, the final forward pipeline is shown in Fig. \ref{fig:real_time_pipeline}. We list all steps in Algorithm \ref{alg:real_time_deploy}. After the training of all model weights, $\mathbf{f}^{\theta_p}$ (Section \ref{sec:uail}),
$\bE_{test}$, $\bE_{train}$, and $\bD_{train}$ (Section \ref{sec:stochastic_transfer}), the whole piepline can be deployed in the testing environment. In each time step, an image $x_{t}$ collected from the sensor mounted on the vehicle in the \textit{Carla} environment under the testing weather condition (\textit{daytime hard rain}) is firstly taken to the encoder of the testing domain $\bE_{test}$ to encode the content code $c_{t}$.
The style codes $\{ s_{tj}' \}_{j=1}^M$ can be encoded from the sampled training domain images $\{ x_{tj}' \}_{j=1}^M$ by the training domain encoder $\bE_{train}$, or directly sampled from a normal distribution. Through the training domain decoder $\bD_{train}$, the original input image $x_t$ is translated to various generated images $\{\tilde{x}_{tj}\}_{j=1}^M$ under different training domain styles. Those generated images will be processed by the pre-trained uncertainty-aware imitation learning policy network $\mathbf{f}^{\theta_p}$. Thus, we get actions and uncertainties corresponding to all the translated images. Among those actions, the one with the lowest uncertainty will be finally deployed to the mobile agent.
Here, we skip the details of the actions ($\tilde{a}_t^\mathbf{a}, \tilde{a}_t^\mathbf{s}, \tilde{a}_t^{\mathbf{b}}$ ), which are all decided by their own uncertainties individually.

\begin{algorithm}[t!]
    \caption{Real-time deployment pipeline}
    \label{alg:real_time_deploy}
    \begin{algorithmic}[0]
    \STATE{
    Pretained model weight $\mathbf{f}^{\theta_p}, \bE_{test}, \bE_{train}$ and $\bD_{train}$. \\
    // At testing time step $t$. \\
    Get the real-time image $x_t \sim \cX_{test}$.\\
    Get the related velocity $v_t$ and high-level command $h_t$. \\
    Encode the content code of $x_t$: $ [ c_t, -\ ] = \bE_{test}(x_t)$.\\}
    \IF {Encode style codes from images in $\mathcal{X}_{train}$, }
      \STATE{
      Sample off-line images $\{ x_{tj}' \}_{j=1}^M \sim \mathcal{X}_{train}$.\\
      Encode style codes $\{ s_{tj}' \}_{j=1}^M$ of the selected style images: $ [ -, s_{tj}' ] = \bE_{train}(x_{tj}')$. \\}
    \ELSE
    \STATE{
    Randomly sample style codes $\{ s_{tj}' \}_{j=1}^M$ from the normal distribution.
    \\}
    \ENDIF
    \STATE{
    Images translation $ \{ \tilde{x}_{tj} \}_{j=1}^M:
    \tilde{x}_{tj} =
     \mathbf{D}_{train} (c_t, s_{tj}')$. \\}
    % \FOR {$j=1$ to $M$}
    \STATE{
    Generate actions through the policy network
    $\{ \tilde{a}_{tj}, \tilde{u}_{tj} \}_{j=1}^M:
    [\tilde{a}_{tj}, \tilde{u}_{tj}] = \mathbf{f}^{\theta_p}(\tilde{x}_{tj}, v_t, h_t)$. \\}
    % \ENDFOR
    \STATE{
    Locate the minimal uncertainty: $j^{*}=\argminE_{j} \{\tilde{u}_{tj}\}_{j=1}^M$. \\
    Output and deploy $\tilde{a}_{tj^{*}}$. \\
    }
\end{algorithmic}
\end{algorithm}

\section{Experiments}
\label{sec:experiments}

\subsection{Model Training}
For the stochastic translation model training, we follow the setup in \cite{Huang_2018_ECCV}\footnote{https://github.com/NVlabs/MUNIT} for $1e6$ steps with batch size as $1$.
As we mentioned before, the training domain $\cX_{train}$ consists of three weather conditions and the testing domain $\cX_{test}$ only contains the images under \textit{daytime hard rain}.
The size of original images in the \textit{Carla} dataset is $200\times88$. To maintain enough information in the content code, they are resized to $256\times256$ for the stochastic image translation.
After that, we get the trained encoders ($\bE_{test}, \bE_{train}$) and decoders ($\bD_{test}, \bD_{train}$),
and they are used in the final forward pipeline, as shown in Section \ref{sec:forward_pipeline}.

The training of uncertainty-aware imitation learning follows the branched structure of conditional imitation learning \cite{Codevilla2018, zhang2018vrsup}. We train all the training domain images (481600 images under three kinds of weather) for 90 epochs with a batch size of 1000.
As in the original setup in \cite{Codevilla2018}, we also try several different network structures for the uncertainty estimation. Experiments show that the current structure processing the feature of the image and the velocity through another four branches outputting uncertainties of actions corresponding to the four high-level commands is the most effective. The code for the imitation learning policy training is available online.\footnote{https://github.com/onlytailei/carla\_cil\_pytorch} We implement all the code through \textit{Pytorch} and all the training is finished by an NVIDIA 1080Ti GPU.

\subsection{Model Evaluation}

\begin{table*}[h!]
\center

\begin{tabular}{lllcclccccc}
\hline
\multicolumn{2}{c}{Policy model}                                                                                                                         &  & \multicolumn{2}{c}{CIL} &  & \multicolumn{5}{c}{UAIL}                                                                                             \\ \hline
\multicolumn{2}{c}{Visual trans. model}                                                                                                                  &  & Direct    & CycleGAN    &  & Direct    & CycleGAN  & \multicolumn{1}{l}{Stoc.-Single.} & \multicolumn{1}{l}{Stoc.-Random.} & Stoc.-Cross          \\ \hline
\multicolumn{2}{c}{Sucess rate(\%)}                                                                                                                      &  & 0.0/-     & 34.7/-      &  & 14.7/16.0 & 44.0/56.0 & 50.1/60.0                         & 54.7/\textbf{64.0}                & \textbf{60.0/64.0}   \\
\multicolumn{2}{c}{Ave. distance to goal travelled(\%)}                                                                                                  &  & 5.2/-     & 55.7/-      &  & 25.8/29.5 & 66.4/78.0 & 73.0/76.3                         & 62.1/67.8                         & \textbf{75.8/79.3}   \\ \hline
\multirow{5}{*}{\begin{tabular}[c]{@{}l@{}}Ave. distance\\ travelled between\\ two infractions\\ in Nav. dynamic\\ (km)\end{tabular}} & Opposite lane    &  & 0.26/-    & 0.83/-      &  & 4.43/6.23 & 1.44/1.77 & 6.91/11.05                        & 11.66/20.67                       & \textbf{12.74/24.58} \\
                                                                                                                                      & Sidewalk         &  & 0.38/-    & 1.29/-      &  & 1.26/1.68 & 2.10/3.16 & 3.72/4.42                         & 3.79/5.17                         & \textbf{5.78/7.70}   \\
                                                                                                                                      & Collision-static &  & 0.16/-    & 0.77/-      &  & 0.37/0.52 & 1.22/1.75 & 2.46/3.16                         & 3.03/6.20                         & \textbf{9.56/23.09}  \\
                                                                                                                                      & Collision-car    &  & 0.27/-    & 0.59/-      &  & 0.56/0.67 & 0.75/0.99 & 1.78/\textbf{2.15}                & 0.61/0.64                         & \textbf{2.04}/2.14   \\
                                                                                                                                      & Collision-ped.   &  & 4.60/-    & 7.29/-      &  & 0.39/0.61 & 8.07/8.85 & 8.17/11.04                        & 5.95/10.34                        & \textbf{9.87/23.58}  \\ \hline
\end{tabular}

\caption{
Quantitative experiments of the \textit{Carla} dynamic navigation benchmark \cite{dosovitskiy2017carla}.
For the imitation learning policy, we compare our \textit{UAIL} with the multi-domain (\textit{CIL}) policy results in \cite{zhang2019vrgoggles}. The proposed uncertainty-aware imitation learning policy overtakes the original \textit{CIL} in all the metrics under both direct deployment (\textit{Direct}) and deterministic transformation through \textit{CycleGAN}. For different visual domain adaptation methods under the \textit{UAIL} policy model, the stochastic methods show much better performances compared with \textit{Direct} and \textit{CycleGAN}. Among them, the proposed \textit{Stocastic-Cross}, considering both the randomization and the directional training styles, achieves the best performance. All the results are shown as the average/max values of the three benchmark trials.
Higher means better for all metrics.
}
\label{tab:carla_metric}
\end{table*}

Finally, we conduct experiments on the \textit{Carla} navigation benchmark mentioned in Section \ref{sec:carla_benchmark}. We compare different strategies both for the policy model and the visual domain transformation methods as follows:

For the policy model, we compare two different setups:
\begin{itemize}
  \item \textbf{CIL}: Map the state to the action without considering the uncertainty as the original conditional imitation learning structure \cite{Codevilla2018}. The output is directly deployed to the vehicle.
  \item \textbf{UAIL}: Take the uncertainty as an output of the network as described in Section \ref{sec:uail}. When using multiple visual inputs, the action with the lowest uncertainty is chosen as the final command.
\end{itemize}

For the visual domain adaptation methods, five strategies are compared:
\begin{itemize}
  \item \textbf{Direct}: Directly deploy the control policy in the testing environment without any visual domain adaptations.
  \item \textbf{CycleGAN}: Transfer the real-time image to a specific training condition through \textit{CycleGAN} \cite{zhu2017unpaired} deterministically.
  \item \textbf{Stochastic-Single}: Directionally transfer the input image to a specific training weather condition based on the style image from the training domain.
  \item \textbf{Stochastic-Random}: Randomly sample three style codes to decode the translated images.
  \item \textbf{Stochastic-Cross}: Directionally transfer the real-time image to all the three training weather conditions based on the style images from the training domain.
\end{itemize}

For the \textit{Stochastic-Single} and \textit{Stochastic-Cross} transfer methods, the style codes are encoded from style images in the training domain. We prepare ten images for each of the training weather conditions. They are randomly sampled from the training dataset under the related weather condition. In each step, \textit{Stochastic-Single} samples one style image from the related weather condition and \textit{Stochastic-Cross} samples three style images from each of the training conditions, respectively.

The \textit{Carla} navigation benchmark consists of four tasks, \textit{Straight}, \textit{One turn}, \textit{Navigation} and \textit{Navigation with dynamic obstacles}, where the vehicle need to finish 25 different navigation routes in each task. Since the first three tasks do not consider any pedestrians or vehicles in the environment, previous methods \cite{Codevilla2018, zhang2019vrgoggles} have achieved considerable generalization results on these tasks. However, in this paper, we mainly consider the most challenging task, \textit{Navigation with dynamic obstacles}, under the testing weather condition.\footnote{The uncertainty-aware policy is a little bit conservative, so we relax the time limit for each trial. However, this does not affect the results of infractions in Table. \ref{tab:carla_metric}}

We run each of the setups three times and show the average/max result through the related benchmarks in Table \ref{tab:carla_metric}. As a deterministic transfer method, we build three transfer models between each training weather condition and the testing weather condition through \textit{CycleGAN}. In each benchmark trial, one specific transfer model is used. The stochastic model can generate various stylized images through a batch operation. However, to achieve such processing through \textit{CycleGAN}, we need to input the real-time image to each of the deterministic transfer models one-by-one, which is both time and resource-consuming. So the three benchmark experiments on \textit{CycleGAN} are under a specific training weather condition for each time. It is the same for the \textit{Stochastic-Single} method, except that the specific training style image is sampled from the prepared subset with a size of ten, as mentioned before.

We do not show the result of combining the \textit{CIL} policy model with the stochastic transfer models because, without uncertainties, there is no reason to choose the specific one among the actions generated through various input images.
The results of \textit{CIL} policy are referred from \cite{zhang2019vrgoggles}, where their \textit{multi-domain} policy is what we mean by \textit{CIL} here. The results in \cite{zhang2019vrgoggles} do not provide the max value of their trials.

Among the different policy models under \textit{Direct} deployment and transformation with \textit{CycleGAN}, our proposed \textit{UAIL} shows great improvements in all of the metrics of the \textit{Carla} navigation benchmark.
In the comparison of different transformation methods, \textit{Stochastic-Cross} shows the best generalization under the testing weather condition.

\begin{figure}[t!]
    \centering
      \includegraphics[width=\columnwidth]{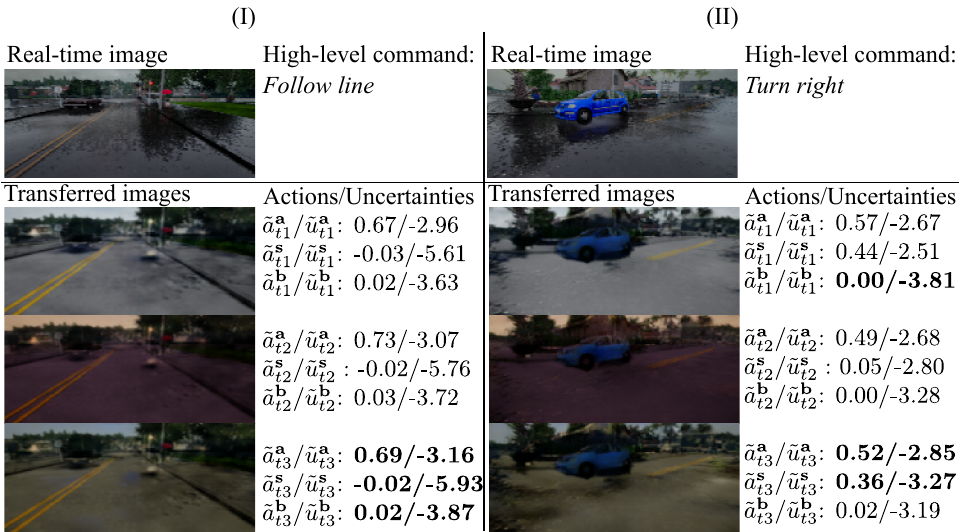}
    \caption{
      Two examples for the proposed pipeline \textit{UAIL} with \textit{Stochastic-Cross}. The final commands with the lowest uncertainties are labelled. For the straight line in (I), which is a relatively clean environment, the uncertainties from all the generated images are almost the same. For the more complex dynamic environment with the turning condition in (II), the chosen steering command $\tilde{a}_{t3}^{\mathbf{s}}$ is much safer. Without an effective steering command, a collision will happen like the $\tilde{a}_{t2}^{\mathbf{s}}$.
    }
    \label{fig:un_example}
\end{figure}

To understand the selection mechanism under our proposed \textit{UAIL} and \textit{Stochastic-Cross} pipeline,
we show two typical uncertainty estimation examples during the testing in Fig. \ref{fig:un_example}.
As shown in Fig. \ref{fig:un_example}-I, the outputs of actions and uncertainties between different translated images are quite close to each other, even though we choose the actions from the last image based on the lowest uncertainty.
Since the straight line scenario is the most common one in the training dataset and the decision is relatively simple to make.
The dynamic and challenging turning scenario in Fig. \ref{fig:un_example}-II is what we are particularly aiming to solve. The second transferred image under \textit{Clear Sunset} outputs a very tiny steering command which could potentially cause a collision with the car in front.

\section{Conclusions}
\label{sec:conclusion}
We proposed a deployment pipeline for a visual-based navigation policy under the \textit{real-to-sim} structure. Through considering the \textit{aleatoric} data uncertainty and the stochastic transformation when translating the testing image back to the training domain, a safer action selection mechanism is constructed for end-to-end driving. Experiments on deploying the pre-trained policy in an unknown extreme weather condition through the \textit{Carla} navigation benchmark show that our proposed pipeline provides a more certain and robust solution.

For future work, finally transferring the trained policy to real-world autonomous driving in a challenging environment would be an exciting next step. Considering the consistency of image streams, like in \cite{zhang2019vrgoggles}, could be another future direction.
Furthermore, this alternative decision-making pipeline may also guide the improvement of the model training for challenging samples. Related model augmentation towards more robust generalization could also be performed.

% \addtolength{\textheight}{-8cm}   % This command serves to balance the column lengths

%%%%%%%%%%%%%%%%%%%%%%%%%%%%%%%%%%%%%%%%%%%%%%%%%%%%%%%%%%%%%%%%%%%%%%%%%%%%%%%%

%%%%%%%%%%%%%%%%%%%%%%%%%%%%%%%%%%%%%%%%%%%%%%%%%%%%%%%%%%%%%%%%%%%%%%%%%%%%%%%%

%%%%%%%%%%%%%%%%%%%%%%%%%%%%%%%%%%%%%%%%%%%%%%%%%%%%%%%%%%%%%%%%%%%%%%%%%%%%%%%%

%\section*{ACKNOWLEDGMENT}

%%%%%%%%%%%%%%%%%%%%%%%%%%%%%%%%%%%%%%%%%%%%%%%%%%%%%%%%%%%%%%%%%%%%%%%%%%%%%%%%

\small
% \footnotesize{tiny}
% \ragged2e
% \spaceskip 0.0em \relax
\bibliographystyle{IEEEtran}
\bibliography{tai19iros}

\begin{thebibliography}{10}
\providecommand{\url}[1]{#1}
\csname url@rmstyle\endcsname
\providecommand{\newblock}{\relax}
\providecommand{\bibinfo}[2]{#2}
\providecommand\BIBentrySTDinterwordspacing{\spaceskip=0pt\relax}
\providecommand\BIBentryALTinterwordstretchfactor{4}
\providecommand\BIBentryALTinterwordspacing{\spaceskip=\fontdimen2\font plus
\BIBentryALTinterwordstretchfactor\fontdimen3\font minus
  \fontdimen4\font\relax}
\providecommand\BIBforeignlanguage[2]{{%
\expandafter\ifx\csname l@#1\endcsname\relax
\typeout{** WARNING: IEEEtran.bst: No hyphenation pattern has been}%
\typeout{** loaded for the language `#1'. Using the pattern for}%
\typeout{** the default language instead.}%
\else
\language=\csname l@#1\endcsname
\fi
#2}}

\bibitem{dosovitskiy2017carla}
A.~Dosovitskiy, G.~Ros, F.~Codevilla, A.~Lopez, and V.~Koltun, ``{CARLA}: {An}
  open urban driving simulator,'' in \emph{CoRL}, vol.~78.\hskip 1em plus 0.5em
  minus 0.4em\relax PMLR, 13--15 Nov 2017, pp. 1--16.

\bibitem{zhang2019vrgoggles}
J.~Zhang, L.~Tai, P.~Yun, Y.~Xiong, M.~Liu, J.~Boedecker, and W.~Burgard,
  ``Vr-goggles for robots: Real-to-sim domain adaptation for visual control,''
  \emph{IEEE Robotics and Automation Letters}, vol.~4, no.~2, pp. 1148--1155,
  April 2019.

\bibitem{Codevilla2018}
F.~Codevilla, M.~Miiller, A.~López, V.~Koltun, and A.~Dosovitskiy,
  ``End-to-end driving via conditional imitation learning,'' in \emph{2018 IEEE
  International Conference on Robotics and Automation (ICRA)}, May 2018, pp.
  1--9.

\bibitem{tai2016deep}
L.~Tai, S.~Li, and M.~Liu, ``A deep-network solution towards model-less
  obstacle avoidance,'' in \emph{2016 IEEE/RSJ International Conference on
  Intelligent Robots and Systems (IROS)}, Oct 2016, pp. 2759--2764.

\bibitem{yang2018eccv}
L.~Yang, X.~Liang, T.~Wang, and E.~Xing, ``Real-to-virtual domain unification
  for end-to-end autonomous driving,'' in \emph{ECCV}.\hskip 1em plus 0.5em
  minus 0.4em\relax Cham: Springer International Publishing, 2018, pp.
  553--570.

\bibitem{Chen_2015_ICCV}
C.~Chen, A.~Seff, A.~Kornhauser, and J.~Xiao, ``Deepdriving: Learning
  affordance for direct perception in autonomous driving,'' in \emph{The IEEE
  International Conference on Computer Vision (ICCV)}, December 2015.

\bibitem{giusti2016machine}
A.~{Giusti}, J.~{Guzzi}, D.~C. {Cireşan}, F.~{He}, J.~P. {Rodríguez},
  F.~{Fontana}, M.~{Faessler}, C.~{Forster}, J.~{Schmidhuber}, G.~D. {Caro},
  D.~{Scaramuzza}, and L.~M. {Gambardella}, ``A machine learning approach to
  visual perception of forest trails for mobile robots,'' \emph{IEEE Robotics
  and Automation Letters}, vol.~1, no.~2, pp. 661--667, July 2016.

\bibitem{zhang2017irosdeep}
J.~Zhang, J.~T. Springenberg, J.~Boedecker, and W.~Burgard, ``Deep
  reinforcement learning with successor features for navigation across similar
  environments,'' in \emph{2017 IEEE/RSJ International Conference on
  Intelligent Robots and Systems (IROS)}, Sep 2017, pp. 2371--2378.

\bibitem{Liang_2018_ECCV}
X.~Liang, T.~Wang, L.~Yang, and E.~Xing, ``Cirl: Controllable imitative
  reinforcement learning for vision-based self-driving,'' in \emph{The European
  Conference on Computer Vision (ECCV)}, September 2018.

\bibitem{tai2018social}
L.~Tai, J.~Zhang, M.~Liu, and W.~Burgard, ``Socially compliant navigation
  through raw depth inputs with generative adversarial imitation learning,'' in
  \emph{2018 IEEE International Conference on Robotics and Automation (ICRA)},
  May 2018, pp. 1111--1117.

\bibitem{gal2016uncertainty}
Y.~Gal, ``Uncertainty in deep learning,'' Ph.D. dissertation, University of
  Cambridge, 2016.

\bibitem{kendall2017uncertainties}
A.~Kendall and Y.~Gal, ``What uncertainties do we need in bayesian deep
  learning for computer vision?'' in \emph{Advances in neural information
  processing systems}, 2017, pp. 5574--5584.

\bibitem{kendall2018multi}
A.~Kendall, Y.~Gal, and R.~Cipolla, ``Multi-task learning using uncertainty to
  weigh losses for scene geometry and semantics,'' in \emph{Proceedings of the
  IEEE Conference on Computer Vision and Pattern Recognition}, 2018, pp.
  7482--7491.

\bibitem{kahn2017uncertainty}
G.~Kahn, A.~Villaflor, V.~Pong, P.~Abbeel, and S.~Levine, ``Uncertainty-aware
  reinforcement learning for collision avoidance,'' \emph{arXiv preprint
  arXiv:1702.01182}, 2017.

\bibitem{lutjens2018safe}
B.~L{\"u}tjens, M.~Everett, and J.~P. How, ``Safe reinforcement learning with
  model uncertainty estimates,'' \emph{arXiv preprint arXiv:1810.08700}, 2018.

\bibitem{henaff2019model}
M.~Henaff, A.~Canziani, and Y.~LeCun, ``Model-predictive policy learning with
  uncertainty regularization for driving in dense traffic,'' \emph{arXiv
  preprint arXiv:1901.02705}, 2019.

\bibitem{choi2018uncertain}
S.~{Choi}, K.~{Lee}, S.~{Lim}, and S.~{Oh}, ``Uncertainty-aware learning from
  demonstration using mixture density networks with sampling-free variance
  modeling,'' in \emph{2018 IEEE International Conference on Robotics and
  Automation (ICRA)}, May 2018, pp. 6915--6922.

\bibitem{pan2017virtual}
X.~Pan, Y.~You, Z.~Wang, and C.~Lu, ``Virtual to real reinforcement learning
  for autonomous driving,'' in \emph{Proceedings of the British Machine Vision
  Conference ({BMVC})}, 2017.

\bibitem{zhu2017unpaired}
J.-Y. Zhu, T.~Park, P.~Isola, and A.~A. Efros, ``Unpaired image-to-image
  translation using cycle-consistent adversarial networks,'' in
  \emph{Proceedings of the IEEE International Conference on Computer Vision},
  2017, pp. 2223--2232.

\bibitem{mueller18corl}
\BIBentryALTinterwordspacing
M.~Mueller, A.~Dosovitskiy, B.~Ghanem, and V.~Koltun, ``Driving policy transfer
  via modularity and abstraction,'' in \emph{Proceedings of The 2nd Conference
  on Robot Learning}, ser. Proceedings of Machine Learning Research,
  A.~Billard, A.~Dragan, J.~Peters, and J.~Morimoto, Eds., vol.~87.\hskip 1em
  plus 0.5em minus 0.4em\relax PMLR, 29--31 Oct 2018, pp. 1--15. [Online].
  Available: \url{http://proceedings.mlr.press/v87/mueller18a.html}
\BIBentrySTDinterwordspacing

\bibitem{Huang_2018_ECCV}
X.~Huang, M.-Y. Liu, S.~Belongie, and J.~Kautz, ``Multimodal unsupervised
  image-to-image translation,'' in \emph{The European Conference on Computer
  Vision (ECCV)}, September 2018.

\bibitem{lee2018diverse}
H.-Y. Lee, H.-Y. Tseng, J.-B. Huang, M.~Singh, and M.-H. Yang, ``Diverse
  image-to-image translation via disentangled representations,'' in
  \emph{Proceedings of the European Conference on Computer Vision (ECCV)},
  2018, pp. 35--51.

\bibitem{almahairi2018augmented}
A.~Almahairi, S.~Rajeswar, A.~Sordoni, P.~Bachman, and A.~Courville,
  ``Augmented cyclegan: Learning many-to-many mappings from unpaired data,''
  \emph{arXiv preprint arXiv:1802.10151}, 2018.

\bibitem{zhang2018vrsup}
\BIBentryALTinterwordspacing
J.~Zhang, L.~Tai, Y.~Xiong, M.~Liu, J.~Boedecker, and W.~Burgard, ``Supplement
  file of {VR-Goggles} for robots: Real-to-sim domain adaptation for visual
  control,'' Tech. Rep., 2018. [Online]. Available:
  \url{https://ram-lab.com/file/tailei/vr\_goggles/supplement.pdf}
\BIBentrySTDinterwordspacing

\end{thebibliography}

\end{document}